# Honey Classification using Hyperspectral Imaging and Machine Learning

Mokhtar A. Al-Awadhi
*Department of Computer Science and IT*
*Dr. Babasaheb Ambedkar Marathwada University*
Aurangabad, India
mokhtar.awadhi@gmail.com

Ratnadeep R. Deshmukh
*Department of Computer Science and IT*
*Dr. Babasaheb Ambedkar Marathwada University*
Aurangabad, India
rrdeshmukh.csit@bamu.ac.in

*Abstract*— In this paper, we propose a machine learning-based method for automatically classifying honey botanical origins. Dataset preparation, feature extraction, and classification are the three main steps of the proposed method. We use a class transformation method in the dataset preparation phase to maximize the separability across classes. The feature extraction phase employs the Linear Discriminant Analysis (LDA) technique for extracting relevant features and reducing the number of dimensions. In the classification phase, we use Support Vector Machines (SVM) and K-Nearest Neighbors (KNN) models to classify the extracted features of honey samples into their botanical origins. We evaluate our system using a standard honey hyperspectral imaging (HSI) dataset. Experimental findings demonstrate that the proposed system produces state-of-the-art results on this dataset, achieving the highest classification accuracy of 95.13% for hyperspectral image-based classification and 92.80% for hyperspectral instance-based classification.

## I. INTRODUCTION

Honey is a natural product with numerous nutritional and health advantages. There are two types of honey: monofloral and polyfloral. Monofloral honey comes from a single botanical source, whereas polyfloral honey comes from a variety of sources. Because monofloral kinds of honey are of better quality, their prices are higher than polyfloral honey kinds. Mislabeling the botanical origin of honey, which affects both consumers and producers, is a common form of honey fraud. As a result, both consumers and producers place a high value on honey verification and classification.

For honey classification, there are two types of analytical methods: traditional and modern [1]. Melissopalynology is a traditional method for classifying honey botanical sources. In this method, the botanical origin of honey is determined through a microscopic investigation of pollen grains in honey. Physicochemical parameter analysis is another traditional method, which is usually used with different machine learning (ML) models for honey authentication. Both melissopalynological and physicochemical parameter analyses are often accurate though, they are time-consuming and tedious. In addition, they require sample preparation and require highly trained professionals.

Modern methods like chromatography, mass spectrometry, and spectroscopy are also used for honey botanical origin classification in combination with ML models. Chromatography and mass spectrometry are accurate, but both are time-consuming procedures that need expensive equipment and arduous sample preparation. On the other hand, spectroscopy in the Near-Infrared (NIR), Mid-Infrared (MIR), and Terahertz (THz) ranges has good accuracy, is relatively quick, and requires little sample preparation. However, spectroscopy requires the use of expensive instruments.

HSI is a technology that combines spectroscopy and spatial imaging, where it has the advantages of both technologies. Combined with ML, HSI provides a non-destructive, fast, inexpensive, and automatic approach for honey botanical origin classification. For instance, the authors in [2] used HSI with SVM, random forest, and radial basis function (RBF) models for classifying five honey botanical origins. The models achieved classification accuracies of 90.52%, 90.92%, and 90.97%, respectively.

The objectives of the present paper are to develop an ML-based method for fast, automatic, and nondestructive classification of honey botanical origins using HSI data. Additionally, the objectives are to improve the performance of the classification methods developed in previous work. The developed method goes over three steps: dataset preparation, feature extraction, and classification.

The dataset used in this research contains hyperspectral (HS) data of 58 different honey types. The honey types are labeled with 21 botanical origins and produced by 11 different brands [3]. The dataset consists of 8700 spectral instances extracted from HS images with a rate of 25 spectral instances per image [4]. Each spectral instance has 128 features that represent various spectral bands. Surface Optic's SOC-710 HS imager was used to acquire HS images of honey samples. The spectral range of the imager is from 400nm to 1000nm with 5nm spectral resolution. HS images have a spatial resolution of 520 X 696 pixels. The HS data of each honey sample were acquired six times. The acquisition attribute in the dataset is used for splitting the dataset into training and test sets and for the cross-validation of the classification models. Figure 1 depicts HS data of honey samples from various botanical origins.

There are some previous works on this dataset given as follows. T. Phillips et al. [5] used a class embodiment autoencoder (CEAE) for feature extraction and a KNN model for classification. Their method achieved a testing accuracy of 90.52% and a cross-validation accuracy of 90.65% on a subset of the dataset comprising 3480 instances. A. Noviyanto et al. [6] obtained a classification accuracy of 91.67% using an SVM classifier. The same author in [7] achieved a testing accuracy of 84.15% on a subset of the dataset using preprocessing, feature extraction, and KNN-based classifier.

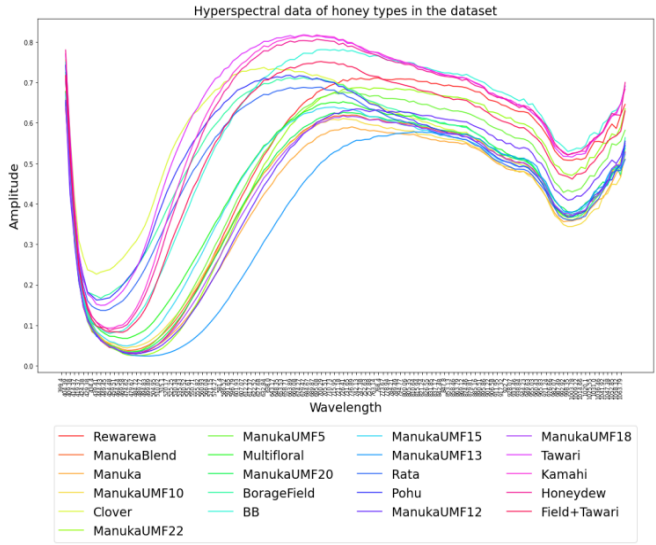

Fig 1 Hyperspectral data of different honey types in the dataset

## II. PROPOSED SYSTEM

The method proposed in this paper consists of three steps: dataset preparation, feature extraction, and classification, as shown in figure 2.

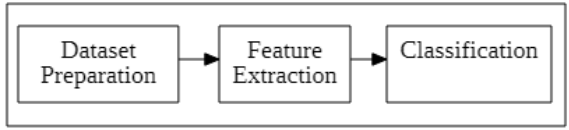

Fig 2 Block diagram of the proposed system

### A. Dataset Preparation

The dataset used in this research contains 21 classes, which represent 21 botanical origins. Each class contains honey samples from different brands. We used the paired t-test to determine the significantly different brands within the same class. Most of the brands were identified as different; consequently, they were treated as separate classes, resulting in 48 total classes in the dataset. The new class labels represent combinations of botanical origins and brands. For instance, honey samples of the class label "Rewarewa" and the brand "C1" are relabeled as "C1_Rewarewa". The performance of the proposed method was evaluated on both original and new class-transformed (CT) datasets.

### B. Feature Extraction

Feature extraction (FE) is essential in many classification problems since it improves the performance of the classification models by extracting only relevant attributes. FE algorithms fall into two types: supervised and unsupervised. Supervised FE algorithms, like LDA, use class labels while extracting features. On the other hand, unsupervised FE algorithms, such as Principal Component Analysis (PCA), do not require class labels for extracting features.

In the present research, we used the LDA algorithm to extract features from the honey HSI dataset. We chose LDA since it achieved good results in similar previous studies. LDA is a linear feature-extraction and dimensionality-reduction algorithm which converts the features into a lower-dimensional space that maximizes the ratio of between-class variation to within-class variance, assuring maximal class separation [8]. There are two different approaches for realizing LDA: class dependent and class independent LDA. In class-dependent LDA, the objective is to maximize the ratio of between-class and within-class for all classes. The goal of the class independent LDA, which is the most common variant, is to maximize the overall between-class variance to the within-class variation. LDA has the benefit of being quick because it just requires the solution of a generalized eigenvalue system [9]. It also works for both binary and multi-class problems, and by adding a quadratic kernel, it can extend to nonlinear LDA.

Feature extraction using LDA is carried out as follows. First, the LDA transformation matrix $W$ is calculated by multiplying the inverse of intra-class matrix $S_W$ by the inter-class matrix $S_B$, as shown in equation 1.

$$W = S_W^{-1} S_B \quad (1)$$

Where:

$$S_w = \sum_j p_j \times (x_j - \mu_j)(x_j - \mu_j)^T \quad (2)$$
$$S_B = \sum_j (\mu_j - \mu) \times (\mu_j - \mu)^T \quad (3)$$
$$\mu_j = \frac{1}{n_j} \sum_{x_i \in \omega_j} x_i \quad (4)$$
$$\mu = \frac{1}{N} \sum_{i=1}^{N} x_i \quad (5)$$

Where $\mu_j$ is the mean value of the jth class, $\mu$ is the mean value of all objects in the dataset, $p_j$ is the prior probability of jth class. Second, the eigenvalues λ and eigenvectors of $W$ are calculated. Third, the eigenvectors that correspond to higher eigenvalues are chosen. Finally, the LDA features are calculated by multiplying the original features matrix by the chosen eigenvectors. The performance of the proposed system was evaluated for a different number of LDA features. To compare the performance of the LDA to that of an unsupervised FE algorithm, we also extracted PCA features in this research.

### C. Classification

The last step of the proposed method is the classification in which we used SVM and KNN models to classify honey botanical origins. We chose the SVM and KNN classifiers because they achieved good performance on the same dataset in similar studies.

The SVM model is an ML classifier that creates an optimum hyperplane in transformed higher-dimensional feature space to separate the classes with minimal errors [10]. The support vector classification algorithm solves the optimization problem given as:

$$\min_{w,b,S} \frac{1}{2} \|w\|^2 + C \sum_{n=1}^{N} S_n$$
$$\text{subject to: } y_n(w \cdot k(x_n) + b) \geq 1 - S_n,$$
$$S_n \geq 0, n = 1, \dots, N \quad (6)$$

Where $\|w\|$ is the length of the weight vector $w$, $b$ is a real constant, $S$ is a slack variables vector, $y_n$ is the class of the $n^{th}$ feature vector $x_n$, $C$ is a regularization factor, $N$ is the number of instances in the training set, and $k(\cdot)$ is a kernel function. We used Python's Scikit-Learn default value of $C$, which is 1, in the experiments. We used two SVM classifiers in the experiments, one with a linear kernel and another with an RBF kernel. Also, we compared the performance of the SVM classifier to the performance of the KNN classifier, as it performed well in previous studies on the same dataset. The value of K used in the experiments was 5, as this value achieved the highest accuracy in previous works.

### D. Performance Evaluation

We used the balanced accuracy (BA) in the present research to evaluate the performance of the proposed classification system. The BA is a robust performance metric, especially for imbalanced datasets [11]. For binary classification problems, the BA is computed as the average of the sensitivity and specificity as follows.

$$BA = (Sensitivity + Specificity)/2 \quad (7)$$

The sensitivity and specificity are performance measures for evaluating the positive and negative class prediction [12]. They can be calculated from the following formula.

$$Sensitivity = TP/(TP + FN) \quad (8)$$
$$Specificity = TN/(TN + FP) \quad (9)$$

Where the true positive (TP) and true negative (TN) are the number of correctly classified positive and negative instances respectively, while the false positive (FP) and false negative (FN) are the number of incorrectly classified positive and negative instances, respectively. The BA for multiclass classification problems is the average of class recalls, where a class recall is equivalent to the class sensitivity.

To compare our results with previous work results, we evaluated the performance of the proposed system for classifying honey botanical origins using cross-validation at 20 folds. In each fold, the training set contains instances from three acquisition numbers, while the test set comprises instances from the remaining acquisition numbers. To simulate the real world, we include all spectral instances representing a hyperspectral image, either in the training set or in the test set.

We assessed the system performance for two classification scenarios: HS instance-based classification and HS image-based classification. In HS instance-based classification, to determine the botanical origin of an HS image, we classify a spectral instance extracted from that HS image. Then, we consider the identified floral origin as the botanical source of that HS image. In HS image-based classification, to determine the botanical origin of an HS image, we classify all the spectral instances extracted from that image. Then, from the classified botanical origins, we consider the most frequent one as the botanical origin of that HS image. Both classification scenarios simulate the real world, where an HS camera captures an HS image of a honey sample. Then, the HS image is segmented, where spectral instances are extracted from the HS image. Next, the LDA algorithm extracts the most discriminating features from the spectral instances. Finally, an ML model classifies the extracted attributes into a particular botanical origin.

### III. RESULTS

#### A. Hyperspectral Instance-based Classification

Table I shows the cross-validation accuracies and standard deviations of the classifiers on the original dataset. The results show the performance of the classifiers using three different feature sets: the original features, PCA features, and LDA features. Table II shows the cross-validation accuracies and standard deviations of the classifiers on the class-transformed dataset. Figures 3 and 4 visualize the results listed in Tables I and II.

Results reveal that the proposed system achieved the highest classification accuracy of 92.80% using an RBF-based SVM classifier, outperforming previous methods. For the original dataset, the KNN classifier achieved the best performance using both the original and reduced features. It achieved classification accuracies of 80.65%, 80.48%, and 77.86% using the original features, PCA features, and LDA features, respectively. For the class-transformed dataset, the KNN classifier also outperformed other classifiers using the original features and PCA features, where achieved classification accuracies of 75.77% and 75.59% using the original features and PCA features, respectively.

TABLE I. PERFORMANCE OF CLASSIFICATION MODELS ON THE ORIGINAL DATASET

| Model | Original features | PCA features | LDA features |
|---|---|---|---|
| KNN | **0.8065 ± 0.0334** | 0.8048 ± 0.0332 | 0.7786 ± 0.0168 |
| SVM (Linear) | 0.3800 ± 0.0180 | 0.3772 ± 0.0183 | 0.7477 ± 0.0167 |
| SVM (RBF) | 0.3396 ± 0.0091 | 0.6304 ± 0.0358 | 0.7645 ± 0.0172 |

TABLE II. PERFORMANCE OF CLASSIFICATION MODELS ON THE CLASS-TRANSFORMED DATASET

| Model | Original features | PCA features | LDA features |
|---|---|---|---|
| KNN | 0.7577 ± 0.0262 | 0.7559 ± 0.0265 | 0.9151 ± 0.0139 |
| SVM (Linear) | 0.6058 ± 0.0257 | 0.6041 ± 0.0254 | 0.9239 ± 0.0140 |
| SVM (RBF) | 0.5102 ± 0.0161 | 0.7011 ± 0.0266 | **0.9280 ± 0.0138** |

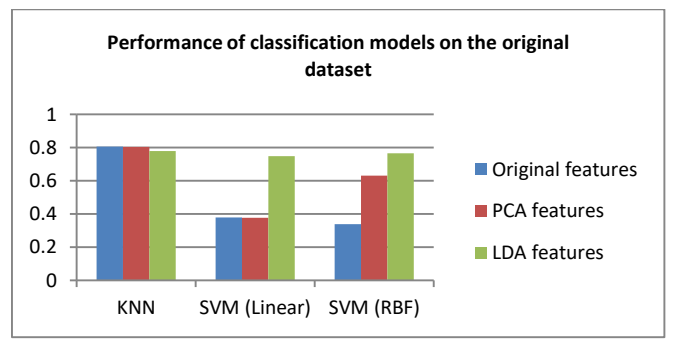

Fig 3 Performance of classification models on the original dataset

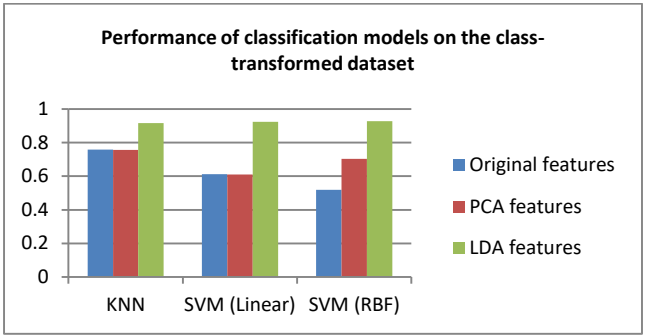

Fig 4 Performance of classification models on the class-transformed dataset

Figure 5 shows the performance of the classifiers on the class-transformed dataset for a different number of LDA features. The graph shows that the classifiers achieved the best performance using the first 15 LDA features.

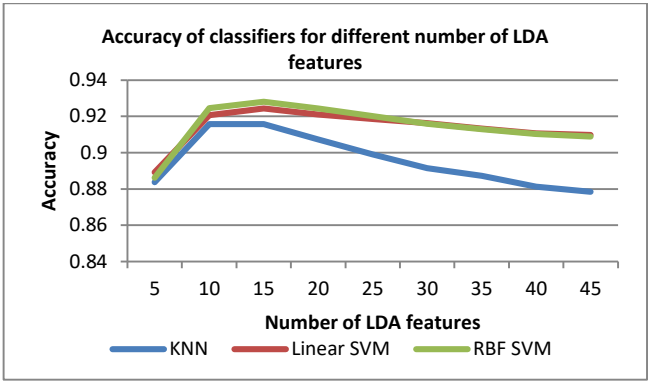

Fig 5 Accuracy of classifiers for different number of LDA features

### B. Hyperspectral Image-based Classification

Table III shows the performance of the classifiers on the class-transformed dataset for hyperspectral image-based classification. We evaluated the performance of the classifiers on the class-transformed dataset using LDA features since this combination achieved the best performance in the instance-based classification scenario. Results reveal that the KNN classifier achieved the highest cross-validation classification accuracy of 95.13% with a standard deviation of 0.0149.

TABLE III. PERFORMANCE OF CLASSIFICATION MODELS ON CLASS-TRANSFORMED DATASET USING LDA FEATURES FOR HYPERSPECTRAL IMAGE BASED CLASSIFICATION

| Model | KNN | SVM (Linear) | SVM (RBF) |
|---|---|---|---|
| Accuracy | **0.9513 ± 0.0149** | 0.9475 ± 0.0143 | 0.9477 ± 0.0172 |

Figure 6 shows a graphical user interface (GUI) of our proposed system. The system accepts HSI data of a honey sample in Comma Separated Values (CSV) format. Then, the system plots the spectral data of the honey sample on the GUI. When a user clicks on the classify button, the system extracts LDA features from the spectral data of the honey sample and inputs them to an SVM classifier. The SVM classifier identifies the botanical origin and brand of the honey sample and displays the result on the GUI. The figure shows a plot of the spectral data of a sample honey type and the identified botanical origin and brand.

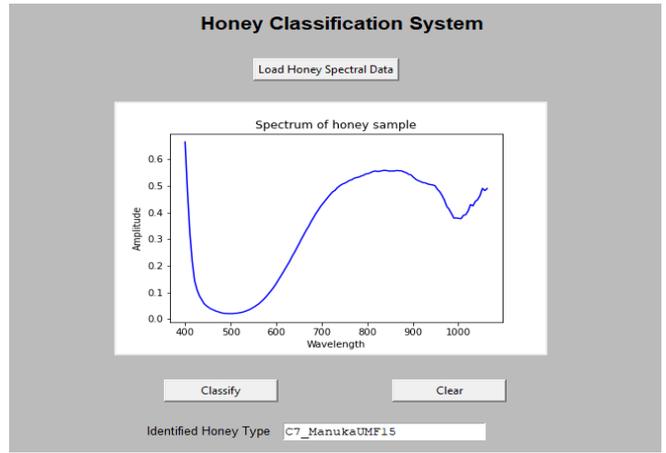

Fig 6 Graphical user interface of the proposed system

### IV. DISCUSSION

Results in Table I and Table II highlight that the classifiers obtain low accuracy using the original features. That is because the features are much correlated, and most of them are irrelevant. Furthermore, FE using PCA did not improve the performance of classifiers on the CT dataset, except RBF SVM. Similarly, there was no significant improvement in the performance of the classifiers on the original dataset using the LDA algorithm. Figure 7 shows the projection of the original dataset on the first and second linear discriminant components. The figure shows how small the inter-class distance is and how high the intra-class variance is. The variation within some classes is high since some brands have significantly different spectral data.

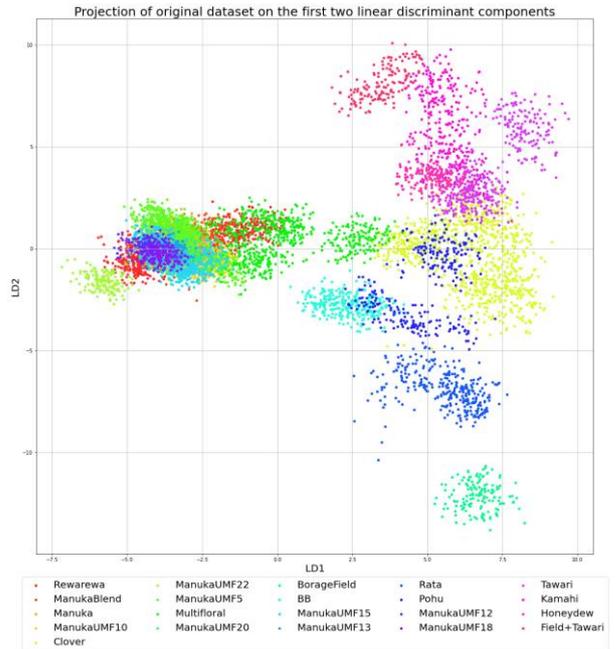

Fig 7 Projection of original dataset on the first and second linear discriminant components

On the other hand, LDA improved the performance of the classifiers on the dataset after class transformation. Figure 8 shows the projection of the class-transformed dataset on the first and second linear discriminant components. The figure shows that the distance between the classes increased while the variance within the classes decreased. Results indicate the effectiveness of the LDA for feature extraction and dimensionality reduction on the class-transformed dataset. The best performance was obtained using the first 15 LDA features since they were the most significant features that explained the majority of the variance in the dataset. Compared to the autoencoder used in previous work, LDA is faster and computationally efficient.

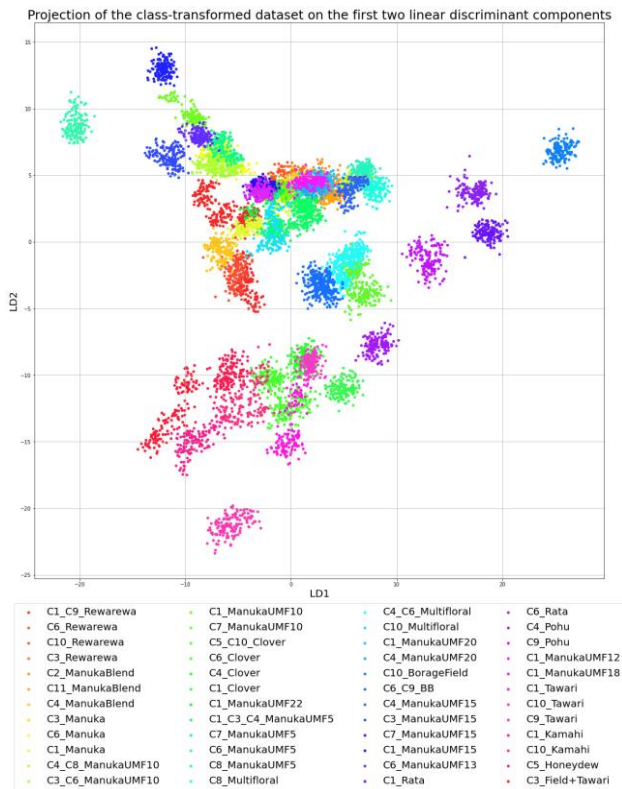

Fig 8 Projection of the class-transformed dataset on the first and second linear discriminant components

Results in Table III show that the proposed system successfully classified around 95% of the hyperspectral images in the dataset. The accuracy increased because of the increased number of instances used in classification. Since the size of the training and test sets has changed, KNN performed slightly better than SVM. The high accuracy obtained in Table III confirms that using all the spectral instances extracted from an HS image for classifying honey botanical origins improved the performance of the classification system. Extracting more spectral instances from the HS images may further increase the classification accuracy.

Table IV shows a comparison of the proposed method performance with the previous work performance. Results indicate that the proposed method outperforms other methods. In addition, results demonstrate that the class transformation, in combination with feature extraction using LDA, improved the performance of the classifiers.

TABLE IV. COMPARISON OF PROPOSED METHOD PERFORMANCE WITH PREVIOUS WORK PERFORMANCE

| Reference | Dataset Size | Feature Extractor | Classifier | Accuracy |
|---|---|---|---|---|
| T. Phillips et al. [5] | 3480 x 128 | CEAE | KNN | 90.65% |
| A. Noviyanto et al. [6] | 8700 x 128 | - | SVM | 91.67% |
| **Proposed Method** | **8700 x 128** | **LDA** | **SVM** | **92.80%** |

## V. CONCLUSION AND FUTURE WORK

Classifying honey botanical origins is significant to consumers and producers as it prevents the fraud of mislabeling botanical sources. Current methods for identifying different honey botanical origins are destructive, expensive, time-consuming, and in some cases require sample preparation. HSI combined with ML provides an automatic, rapid, economical, and nondestructive alternative method for classifying honey botanical origins.

A system for automatically classifying honey botanical sources was proposed in this research. Dataset preparation, feature extraction, and classification are all part of the proposed system. The performance of the proposed method was evaluated using a standard honey hyperspectral image dataset. In the dataset preparation stage, a statistical t-test was employed to determine significantly different brands within the same classes. Thus, various brands were treated as separate classes. In the feature extraction stage, the LDA algorithm was used to extract the most relevant features. In the classification stage, SVM and KNN classifiers were applied to discriminate between the honey botanical origins. Findings demonstrate the capability of the HSI technology with ML in providing quick, automatic, and nondestructive honey classification. Our next step is applying our method to other honey spectral datasets.

REFERENCES

[1] N. Ling Chin and K. Sowndhararajan, "A Review on Analytical Methods for Honey Classification, Identification and Authentication," Honey Anal. - New Adv. Challenges, 2020, doi: 10.5772/intechopen.90232.

[2] S. Minaei et al., "VIS/NIR imaging application for honey floral origin determination," Infrared Phys. Technol., vol. 86, pp. 218–225, 2017, doi: 10.1016/j.infrared.2017.09.001.

[3] T. Phillips, A. Noviyanto, and W. Abdulla, "Hyperspectral Imaging honey dataset," pp. 3–6, 2020, [Online]. Available: https://figshare.com/s/25afe30ff531b8f1e65f.

[4] A. Noviyanto and W. H. Abdulla, "Segmentation and calibration of hyperspectral imaging for honey analysis," Comput. Electron. Agric., vol. 159, no. January, pp. 129–139, 2019, doi: 10.1016/j.compag.2019.02.006.

[5] T. Phillips and W. Abdulla, "Class Embodiment Autoencoder (CEAE) for classifying the botanical origins of honey," Int. Conf. Image Vis. Comput. New Zeal., vol. 2019-Decem, 2019, doi: 10.1109/IVCNZ48456.2019.8961004.

[6] A. Noviyanto and W. H. Abdulla, "Honey botanical origin classification using hyperspectral imaging and machine learning," J. Food Eng., vol. 265, no. January 2019, p. 109684, 2020, doi: 10.1016/j.jfoodeng.2019.109684.

[7] A. Noviyanto, "Honey botanical origin classification using hyperspectral imaging and machine learning," Ph.D. dissertation, The University of Auckland, 2018.

[8] A. Tharwat, T. Gaber, A. Ibrahim, and A. E. Hassanien, "Linear discriminant analysis: A detailed tutorial," AI Commun., vol. 30, no. 2, pp. 169–190, 2017, doi: 10.3233/AIC-170729.

[9] P. Xanthopoulos, P. Pardalos, and T. Trafalis, Robust Data Mining. 2013.

[10] C. Cortes and V. Vapnik, "Support-vector networks," Mach. Learn., vol. 20, no. 3, pp. 273–297, 1995, doi: 10.1007/BF00994018.

[11] K. H. Brodersen, C. S. Ong, K. E. Stephan, and J. M. Buhmann, "The balanced accuracy and its posterior distribution," Proc. - Int. Conf. Pattern Recognit., pp. 3121–3124, 2010, doi: 10.1109/ICPR.2010.764.

[12] C. Sammut and G. I. Webb, "Encyclopedia of machine learning and data mining". Springer, 2017.